\documentclass[runningheads]{llncs}
\usepackage[T1]{fontenc}
\usepackage{graphicx}
\usepackage{booktabs}
\usepackage{multirow}
\usepackage{tabularx}
\usepackage{CJKutf8}
\usepackage{makecell}
\usepackage{hhline}
\usepackage{xcolor}
\usepackage{float}
\usepackage{hyperref}
\hypersetup{colorlinks=true, allcolors=blue}
\usepackage{marvosym}
\begin{document}
\title{MCCD: A Multi-Attribute Chinese Calligraphy Character Dataset Annotated with Script Styles, Dynasties, and Calligraphers}
\titlerunning{MCCD: A Multi-Attribute Chinese Calligraphy Character Dataset}
%
%
\author{Yixin Zhao\orcidID{0009-0002-1101-4910} \and
Yuyi Zhang\orcidID{0009-0009-2721-1741} \and
Lianwen Jin\textsuperscript{(\Letter)}\orcidID{0000-0002-5456-0957}}

\authorrunning{Y. Zhao et al.}
\institute{South China University of Technology, Guangzhou, China \\
\email{yixin\_zhao01@126.com, yuyi.zhang11@foxmail.com, eelwjin@scut.edu.cn}}
\maketitle             
\begin{abstract}
Calligraphy represents an invaluable cultural heritage of Chinese civilization. Research on the attribute information of calligraphy, such as styles, dynasties, and calligraphers, holds significant cultural and historical value. However, the styles of Chinese calligraphy characters have evolved dramatically through different dynasties and the unique touches of calligraphers, making it highly challenging to accurately recognize these different characters and their attributes. Furthermore, existing calligraphic datasets are extremely scarce, and most provide only character-level annotations without additional attribute information. This limitation has significantly hindered the in-depth study of Chinese calligraphy. To fill this gap, we present a novel \underline{M}ulti-Attribute \underline{C}hinese \underline{C}alligraphy Character \underline{D}ataset (MCCD). The dataset encompasses 7,765 categories with a total of 329,715 isolated image samples of Chinese calligraphy characters, and three additional subsets were extracted based on the attribute labeling of the three types of script styles (10 types), dynasties (15 periods) and calligraphers (142 individuals). The rich multi-attribute annotations render MCCD well-suited diverse research tasks, including calligraphic character recognition, writer identification, and evolutionary studies of Chinese characters. We establish benchmark performance through single-task and multi-task recognition experiments across MCCD and all of its subsets. The experimental results demonstrate that the complexity of the stroke structure of the calligraphic characters, and the interplay between their different attributes, leading to a substantial increase in the difficulty of accurate recognition. MCCD not only fills a void in the availability of detailed calligraphy datasets but also provides valuable resources for advancing research in Chinese calligraphy and fostering advancements in multiple fields. The dataset is available at \url{https://github.com/SCUT-DLVCLab/MCCD}.
\end{abstract}
\section{Introduction}
Chinese calligraphy is an ancient and important art form, and is an important part of traditional Chinese culture. With the change of dynasties and the continuous evolution of Chinese civilization, Chinese calligraphy fonts have undergone significant changes, forming a variety of styles. Different calligraphic styles have both morphological connections and obvious structural differences between them. At the same time, the embellishments of different calligraphers' writing habits have made the stroke order and style of calligraphic characters varied and distinctive, which makes it extremely difficult to recognize calligraphic characters and their attributes, such as calligraphy style, dynastic origin and calligrapher. Recently, the preservation and inheritance of Chinese calligraphy have been receiving increasing attention \cite{pengcheng2017chinese,li2022swordnet,Yang2023Open}. However, most existing Chinese character datasets \cite{zhang2009hcl2000,su2006hit,zhou2010hit,jin2011scut,liu2011casia} consist mainly of handwritten simplified Chinese characters and, like many current historical Chinese character datasets \cite{ma2020joint,xu2019casia,shi2025large}, contain only character labels at the character level, which is limited to be used for character recognition research. But the complex styles and variable stroke structures of Chinese calligraphy characters render the direct migration of simplified Chinese character recognition methods usually ineffective. 

\begin{figure}[t]
\centering
\includegraphics[width=1\linewidth]{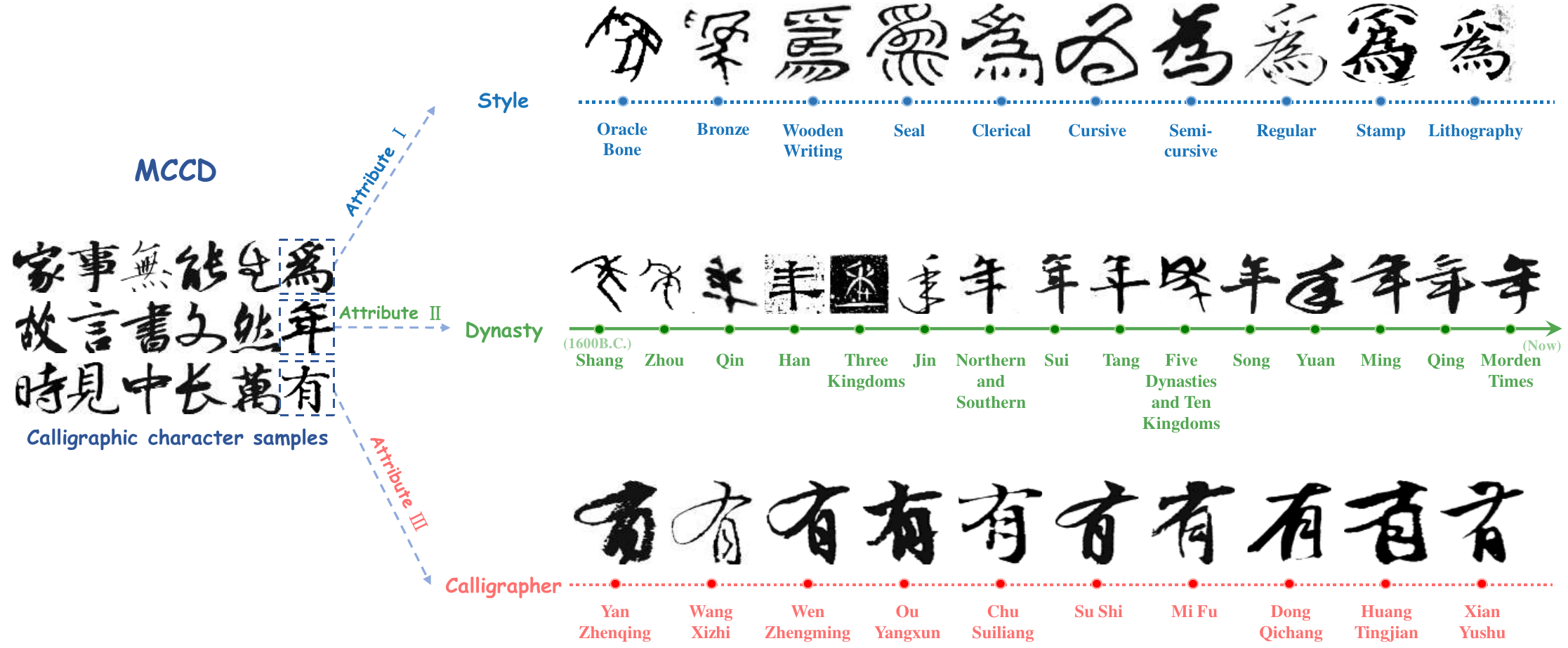}
\caption{An overview of the proposed MCCD. Some samples of Chinese calligraphy characters in MCCD are shown, as well as samples of different script styles of the character ‘\begin{CJK*}{UTF8}{gbsn}为\end{CJK*}’, samples of different dynasties of the character ‘\begin{CJK*}{UTF8}{gbsn}年\end{CJK*}’, and samples of different calligraphers of the character ‘\begin{CJK*}{UTF8}{gbsn}有\end{CJK*}’.} \label{DS}
\end{figure}

While some datasets \cite{wang2024open,guan2024open,wang2024puzzle} attempt to capture the historical evolution of Chinese characters, they are solely confined to a single calligraphic style or a specific historical period and do not focus on the attributes of Chinese characters from other perspectives. The process of collecting and labeling calligraphic corpora is a time-consuming and labor-intensive task, thus leading to little success in the existing construction of the calligraphic characters datasets. The lack of a systematic open-source dataset of Chinese calligraphy characters and benchmark evaluation metrics further limits research on critical aspects of calligraphy research, including script style analysis, calligrapher identification, and character and style evolution.

To bridge this gap and advance calligraphic research in the digital era, we present a novel \underline{M}ulti-Attribute \underline{C}hinese \underline{C}alligraphy Character \underline{D}ataset (MCCD). This dataset is unique in its comprehensive coverage of four annotation dimensions of information: characters, script styles, dynasties, and calligraphers, some of the samples in the dataset are shown in Fig.~\ref{DS}. This multi-attribute annotated dataset not only facilitates the digitization of cultural heritage, but also provides comprehensive data support for the study of the characteristics between artistic expression and cultural context of Chinese calligraphy. Our contributions are summarized as follows.

\begin{itemize}
    \item \textbf{Extensive Multi-Attribute Collection:} MCCD dataset presents a meticulously curated collection of nearly 330,000 calligraphic character images, ensuring a comprehensive diversity of annotation categories for all characters and their attributes (style, dynasty, and calligrapher).
    \item \textbf{Multi-Attribute Subset Construction:} MCCD contains labels for 7,765 categories of characters, in addition to which three additional subsets are extracted from the dataset according to the attribute annotations for each character, including 10 styles of calligraphy, 15 major historical dynasties and 142 famous calligraphers, with the aim of optimizing task-specific utilization of the attribute information.
    \item \textbf{Benchmark Establishment:} We established benchmark performance metrics for single-task recognition (character and each attribute independently) and multi-task recognition (character combined with other attributes simultaneously) experiments using MCCD and all its subsets.
\end{itemize}

\section{Related Works}
With the growing prominence of Chinese character recognition research, Chinese character datasets have evolved to meet diverse research and application demands. Existing datasets can be broadly categorized into three groups: modern handwritten Chinese character datasets, historical and ancient script datasets and large-scale comprehensive datasets.

\textbf{Modern Handwritten Chinese Character Datasets:} Modern handwritten Chinese character datasets primarily focus on commonly used simplified characters in contemporary contexts. For example, HCL2000 \cite{zhang2009hcl2000} includes 3,755 frequently used simplified Chinese characters written by 1,000 distinct writers, along with sub-databases containing writer demographics. HIT-MW \cite{su2006hit} comprises 853 handwritten document samples from 780 writers, yielding 186,444 segmented characters (3,041 categories), including letters and punctuation. HIT-OR3C \cite{zhou2010hit} features 832,650 samples of 6,825 categories (covering common characters, letters, and digits) from 122 writers, supplemented by 77,168 isolated characters (2,442 categories) extracted from 10 documents authored by 20 writers. SCUT-COUCH2009 \cite{jin2011scut}, an expanded version of SCUT-COUCH2008 \cite{li2008scut}, contains over 3.6 million character samples across 11 subsets contributed by more than 190 writers. CASIA-HWDB \cite{liu2011casia} provides approximately 3.9 million isolated character samples (7,356 categories) written by 1,020 writers.

\textbf{Historical and Ancient Script Datasets:} Historical and ancient script datasets encompass characters collected from archival documents and ancient manuscripts, capturing diverse script styles across historical periods. Representative examples include: MTHv2 \cite{ma2020joint} (expanded from TKH-MTH \cite{yang2018dense}) contains 2,200 text images from two historical Chinese archives, covering 6,733 character categories with over 1 million samples. CASIA-AHCDB \cite{xu2019casia} comprises 2.2 million annotated handwritten character samples (10,350 categories) from ancient Chinese manuscripts. HisDoc1B \cite{shi2025large}, a large-scale historical document dataset, includes over 1 billion characters across 306,125 categories within document images. HUST-OBC \cite{wang2024open} focuses on oracle bone script, containing 140,053 samples of 1,588 deciphered and 9,411 undeciphered characters. EVOBC \cite{guan2024open} collects 229,170 character images (13,714 categories) spanning six major evolutionary stages of Chinese characters. ACCP \cite{wang2024puzzle} extends HUST-OBC \cite{wang2024open} and EVOBC \cite{guan2024open} by incorporating characters from the Kangxi Dictionary, totaling more than 340,000 character images across nearly 90,000 categories.

\textbf{Large-Scale Comprehensive Datasets:} Significantly surpassing previous works, MegaHan97K \cite{zhang2025megahan97k} introduces a massive Chinese character dataset with 97,455 categories, featuring diverse handwritten, historical, and synthetic subsets while being the first to support the comprehensive GB18030-2022\footnote{https://en.wikipedia.org/wiki/GB\_18030} standard.

Existing Chinese character datasets, though valuable, lack essential calligraphic metadata such as script style, dynasty, and calligrapher. This deficiency limits their utility for in-depth calligraphic studies, as they fail to capture the art form's stylistic, historical, and cultural dimensions. To address this gap, the development of a multi-attribute annotated dataset is imperative. This would not only facilitate research into the rich heritage of calligraphy but also provide fine-grained data to advance deep learning-based recognition models.

\section{MCCD Dataset}
\subsection{Dataset Create Processs}
The aim of constructing MCCD is to provide a comprehensive dataset of calligraphic Chinese characters, incorporating as much calligraphy attribute information as possible, such as dynasties, styles, and calligraphers, with a view to further promoting research on calligraphy. Guided by this vision, we made extensive efforts to collect high-quality calligraphic character images and their associated attribute information.

\textbf{(1) Data Collection.} We investigated numerous online resources on Chinese calligraphy characters and ultimately selected two authoritative calligraphy websites: ‘\textit{ZiTong}' (\begin{CJK*}{UTF8}{gbsn}字统\end{CJK*})\footnote{https://zi.tools/} and ‘
\textit{ShuFaTuJi}' (\begin{CJK*}{UTF8}{gbsn}书法图集\end{CJK*})\footnote{https://www.sftj.com/}. These platforms provide high-resolution authentic calligraphic character images accompanied by rich metadata including script styles, calligraphers, and historical dynasties, which significantly facilitates the construction of a comprehensive calligraphy dataset. Through our well-designed web crawler tool, we maximized data acquisition from both websites while ensuring precise image-label correspondence. Each sample is annotated with calligraphic character categories and their attributes such as script style, dynasty and calligrapher. For samples missing certain attributes, the missing information is denoted as ‘null’. 

\textbf{(2) Data Cleaning.} After the collection, we merged the data from both sources. However, a significant portion of the collected images contained two duplicate characters in a single image or had garbled and incorrect stock annotations. To address these issues, we manually inspected the images to weed out images with such conditions. In addition, we also removed blurred images. Two annotators each spent about 50 hours validating the dataset in this process.

\textbf{(3) Dataset Construction.} To make the most efficient use of the information in each attribute, we filtered out three subsets of categories from the cleaned data, each based on a specific type of attribute. The structure of the entire MCCD dataset is as follows: 
\begin{itemize}
    \item \textbf{MCCD:} A total set comprises the complete collection of all image samples. As all samples contain corresponding character information, this dataset is annotated with character categories.
    \item \textbf{MCCD-Style:} A subset annotated based on the calligraphy style attributes that are present in the samples, which can be used to explore the differences in calligraphic styles and the evolutionary relationship between Chinese characters and script styles, etc.
    \item \textbf{MCCD-Dynasty:} A subset annotated based on the historical dynasty attributes that are present in the samples,which can be used to explore the historical evolution of Chinese calligraphy characters. A small number of dynasties that were close in time and had little change in calligraphic style were merged to preserve the main periods of Chinese history.
    \item \textbf{MCCD-Calligrapher:} A subset annotated based on the individual calligrapher attributes that are present in the samples,which can be used in the field of writer identification for exploring stylistic differences between calligraphers. Well-known calligraphers were prioritized for retention to ensure adequate samples and representative personalities.
\end{itemize} This construction method ensure that the dataset focuses on analyzing high-quality and representative samples. Furthermore, the structured annotation categorization facilitates accurate and efficient utilization of each attribute type for various research applications.

\subsection{Data Format}
\begin{figure}[t]
\centering
\includegraphics[width=0.6\linewidth]{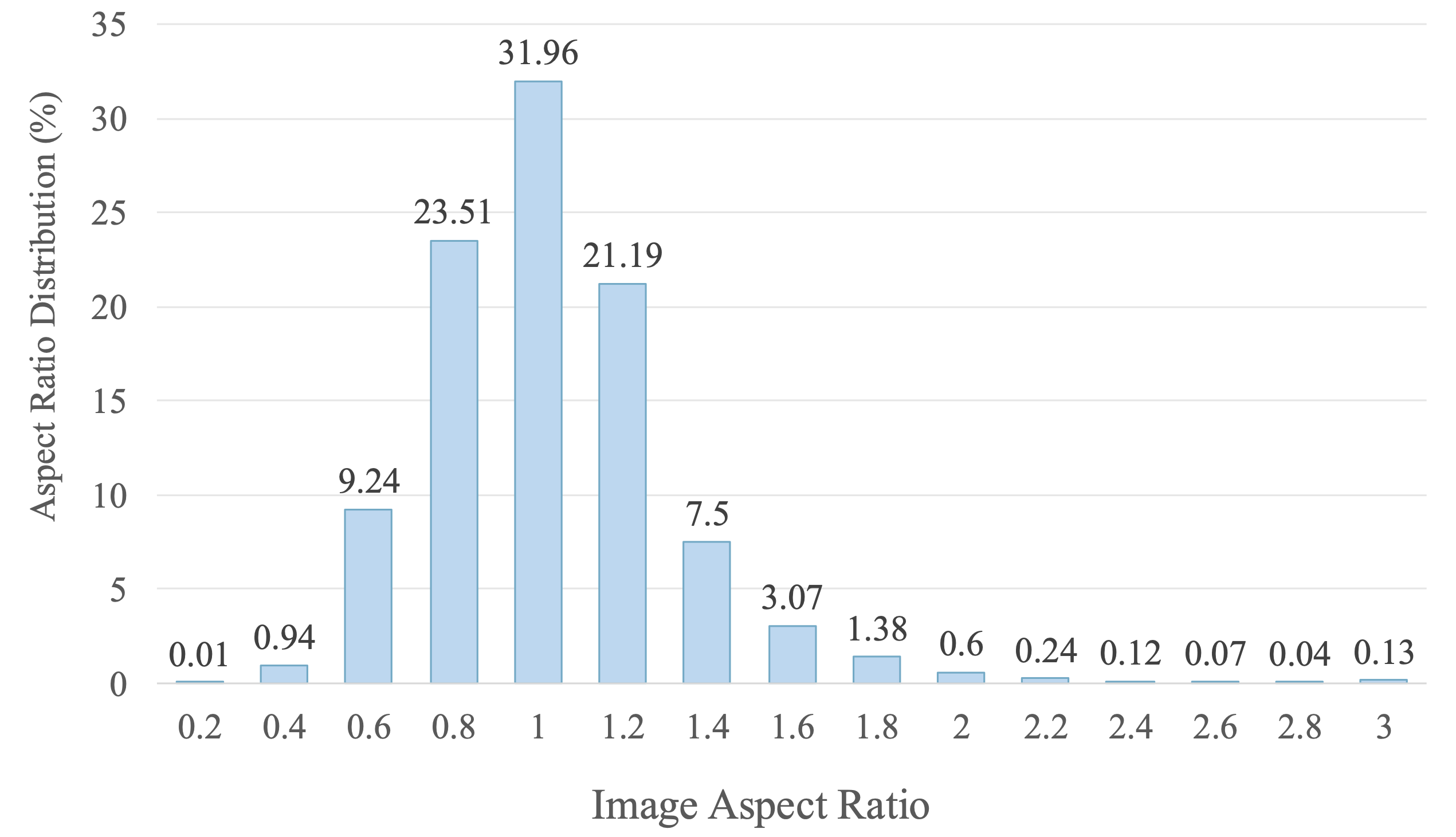}
\caption{The aspect ratio of the images} \label{aspect_ratio}
\end{figure}

As shown in Table~\ref{tab:datasource}, we collected a total of 329,715 Chinese calligraphy character samples from the two sources: ‘\textit{ZiTong}’ contributed 258,830 samples, while ‘\textit{ShuFaTuJi}’ provided 70,885 samples. All images in the dataset are isolated-character crops extracted directly from original calligraphic works. Table~\ref{tab:zhi} demonstrates the samples of ‘\begin{CJK*}{UTF8}{gbsn}之\end{CJK*}’ along with its corresponding multi-attribute labels. Due to variations in stroke lengths and cropping ratios across different sources, the image dimensions differ slightly. As shown in Fig.~\ref{aspect_ratio}, image aspect ratios are primarily concentrated between 0.6 and 1.2 (mean=0.9241, standard deviation=0.2988), spanning a full range from under 0.2 to over 3.0. All samples retain their original colors and are stored in PNG format.

\renewcommand{\arraystretch}{1.25}
\setlength{\tabcolsep}{8pt}
\begin{table}[t]
  \centering
  \caption{Statistics on data collection from different sources.}
    \resizebox{0.48\textwidth}{!}{
    \begin{tabular}{ccc}
    \toprule
    Data Source & \# Samples & \# Total \\
    \hline
    \textit{‘ZiTong’} & 258,830 & \multirow{2}[1]{*}{329,715} \\
    \cmidrule{1-2}
    \textit{‘ShuFaTuJi’} & 70,885 &  \\
    \bottomrule
    \end{tabular}}
  \label{tab:datasource}%
\end{table}%

\renewcommand{\arraystretch}{1.7}
\setlength{\tabcolsep}{8pt}
\begin{table}[t]
  \centering
  \caption{Examples of multi-attribute labels of ‘\begin{CJK*}{UTF8}{gbsn}之\end{CJK*}’ in the MCCD dataset.}
  \resizebox{0.9\textwidth}{!}{
   \begin{tabular}{cccccc}
    \hline
    \textbf{Style} & Oracle Bone & Bronze & Stamp & Lithography & Wooden  \\
    \hline
    
   \textbf{Dynasty} & Shang & Zhou & Song & Han & Qin \\
    \hline
    \textbf{Calligrapher} & \textbackslash{} & \textbackslash{} & \textbackslash{} & \textbackslash{} & Li Ye \\
    \hline
    \makecell{\textbf{Image}} &
    \makecell{\includegraphics[width=0.05\linewidth]{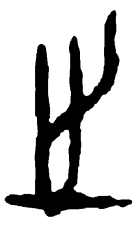}} &
    \makecell{\includegraphics[width=0.09\linewidth]{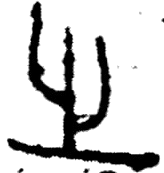}} &
    \makecell{\includegraphics[width=0.07\linewidth]{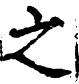}} &
    \makecell{\includegraphics[width=0.08\linewidth]{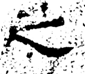}} &
    \makecell{\includegraphics[width=0.08\linewidth]{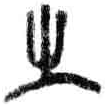}}  \\
    \hline \hline
    \textbf{Style} & Cursive & Semi-Cursive & Seal  & Clerical & Regular  \\
    \hline
    \textbf{Dynasty} & Tang  & Song  & Yuan  & Qing  & Sui \\
    \hline
    \textbf{Calligrapher} & Huai Su & Mi Fu  & Wu Rui & Deng Shiru & Zhi Yong \\
    \hline
    \makecell{\textbf{Image}} &
    \makecell{\includegraphics[width=0.06\linewidth]{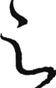}} &
    \makecell{\includegraphics[width=0.08\linewidth]{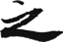}} &
    \makecell{\includegraphics[width=0.08\linewidth]{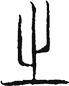}} &
    \makecell{\includegraphics[width=0.08\linewidth]{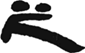}} &
    \makecell{\includegraphics[width=0.08\linewidth]{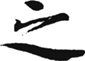}}  \\
    \hline
    \end{tabular}}
  \label{tab:zhi}%
\end{table}%

The final reconstructed dataset is shown in Table~\ref{tab:restructured_dataset} which consists of the total MCCD dataset of 7765 Chinese character annotations and its three subsets, MCCD-Style for 10 calligraphic styles, MCCD-Dynasty for 15 major historical periods, and MCCD-Calligrapher for 142 famous calligraphers. For each set, data under the same category folder are split into training and testing sets at a 7:3 ratio. To enhance data retrieval efficiency, all training and testing sets are also converted into the LMDB format (binary form).

\subsection{Dataset Structure and Statistic}
\begin{figure}
\centering\includegraphics[width=0.45\linewidth]{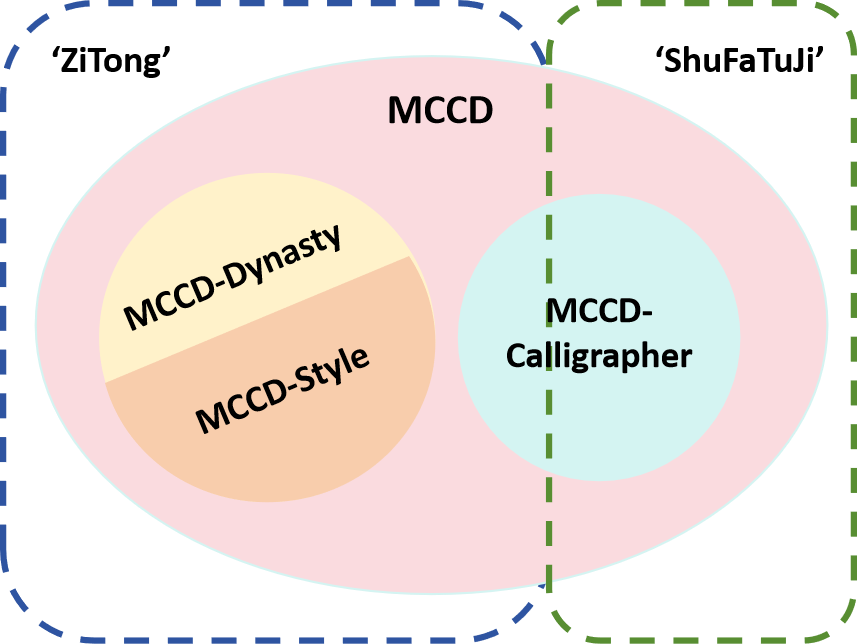}
\caption{Venn diagrams of the relationship between MCCD and its three subsets.}\label{ven}
\end{figure}

While we extracted three subsets from MCCD based on attribute types, it is important to note that there is some overlap in the samples between the subsets. Sample selection for subsets depends on two criteria: the presence of relevant attribute labels and the sample distribution across categories. The Venn diagram in Fig.~\ref{ven} illustrates the relationship between MCCD and its three subsets. The samples from both of the two sources have character labels, while only the data from the \textit{‘ZiTong’} contains complete information about the attributes of the calligraphic style and dynasty, and the attribute of the calligrapher need to be finely selected from both of the two sources. The following sections provide detailed descriptions of the entire dataset.

\renewcommand{\arraystretch}{1.25}
\setlength{\tabcolsep}{8pt}
\begin{table}[t]
  \centering
  \caption{Statistics of training and testing datasets of MCCD.}
    \resizebox{0.8\textwidth}{!}{
    \begin{tabular}{ccccc}
     \toprule
    \multicolumn{2}{c}{Dataset} & \# Samples & \# Categories & \# Total \\
     \hline
    \multirow{2}[1]{*}{MCCD} & train & 234,225 & 7765  & \multirow{2}[1]{*}{329,715} \\
    \cmidrule(lr){2-4}
          & test  & 95,460 & 7765  &  \\
     \hline
    \multirow{2}[1]{*}{MCCD-Style} & train & 181,186 & 10    & \multirow{2}[1]{*}{258,830} \\
    \cmidrule(lr){2-4}
          & test  & 77,644 & 10    &  \\
     \hline
    \multirow{2}[1]{*}{MCCD-Dynasty} & train & 1,811,187 & 15    & \multirow{2}[1]{*}{258,830} \\
    \cmidrule(lr){2-4}
          & test  & 77,643 & 15    &  \\
     \hline
    \multirow{2}[1]{*}{MCCD-Calligrapher} & train & 64,544 & 142   & \multirow{2}[1]{*}{92,122} \\
    \cmidrule(lr){2-4}
          & test  & 27,578 & 142   &  \\
     \bottomrule
    \end{tabular}}
  \label{tab:restructured_dataset}%
\end{table}%

\begin{figure}
\centering
\includegraphics[width=1\linewidth]{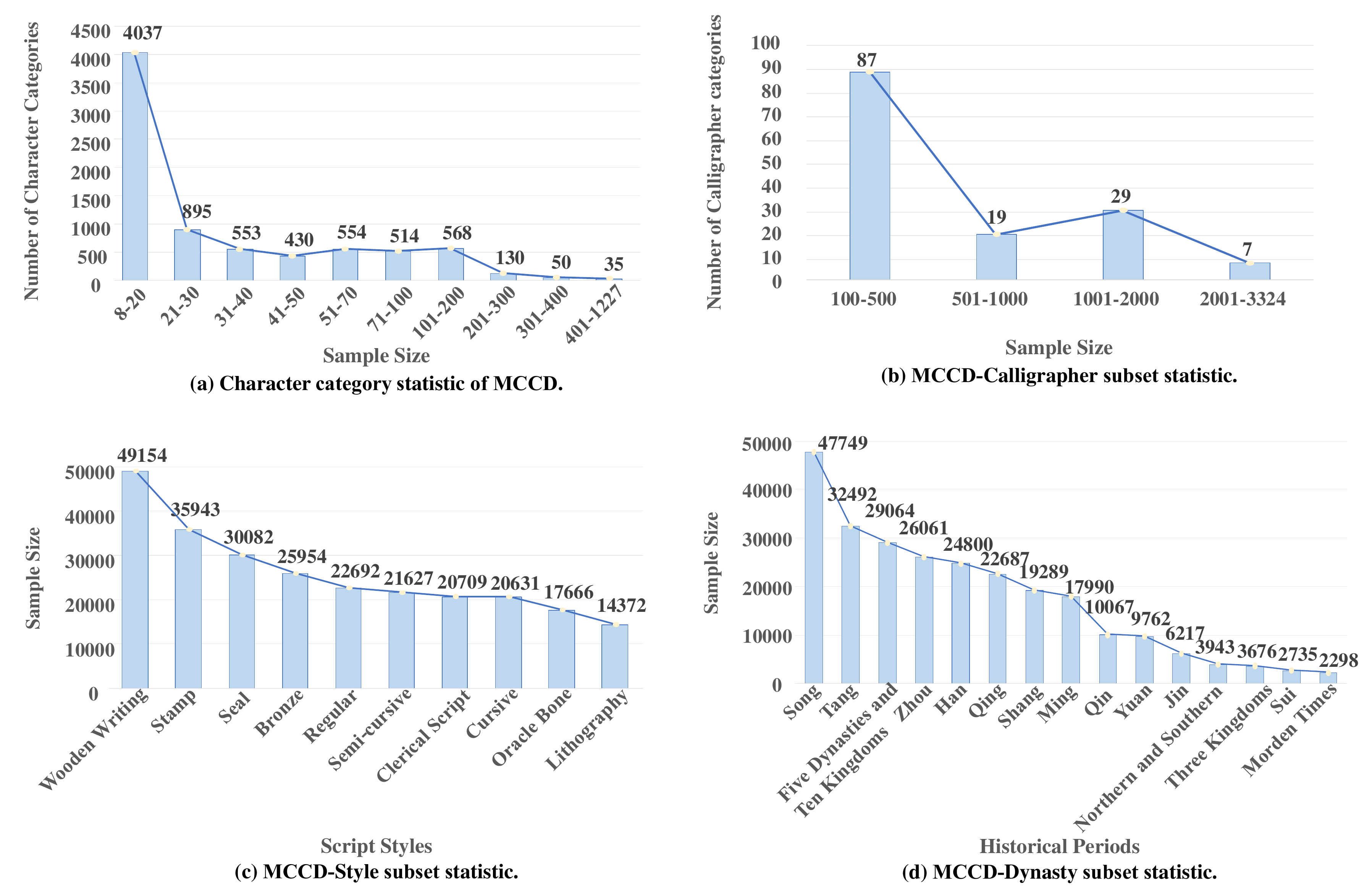}
\caption{Statistical overview of MCCD. As for (a) and (b), the horizontal axis represents the range of sample sizes, and the vertical axis indicates the number of the categories within each range. As for (c) and (d), the horizontal axis represents the categories, and the vertical axis indicates the sample size for each category.}\label{AS}
\vspace{-0.3cm}
\end{figure}

\textbf{MCCD:} A total set consists of all 329,715 samples collected from two sources, all of which have character labels, for a total of 7,765 distinct Chinese character categories. As shown in Fig.~\ref{AS}a, each character category contains no fewer than 8 samples, with approximately 48\% of the categories containing over 20 images. This dataset is divided into 234,225 training samples and 95,460 testing samples.

\textbf{MCCD-Calligrapher:} The subset of calligrapher attributes was carefully selected from all the samples in the MCCD, retaining only renowned calligraphers who were widely recognized. Each category contains no fewer than 100 images, with 36 calligraphers having over 1,000 samples, as shown in Fig.~\ref{AS}b. This subset comprises 92,122 samples covering 142 calligrapher categories. We sampled data with too large a gap in sample size to ensure an average number of samples in each category and to reduce the influence of differences in the number of samples in different categories, thus making the assessment of the recognition performance of calligraphers more reliable. This subset is divided into 64,544 training samples and 27,578 testing samples.

\textbf{MCCD-Style:} The subset of calligraphic style attributes in MCCD contains 258,830 samples all from \textit{‘ZiTong’}. In addition to the five common calligraphic styles of Seal Script, Clerical Script, Cursive Script, Semi-Cursive Script, and Regular Script, five special calligraphic styles of Oracle Bone Inscriptions, Bronze Inscriptions, Wooden Writing Script, Stamp Script, and Lithography have been added to the dataset, making up a total of 10 calligraphic style categories. As depicted in Fig.~\ref{AS}c, each style category contains no fewer than 10,000 images, with 80\% of the categories exceeding 20,000 images, ensuring sufficient data representation across all styles. This subset is partitioned into 181,186 training samples and 77,644 testing samples.

\textbf{MCCD-Dynasty:} The subset of historical dynasty attributes in MCCD is composed of 258,830 samples exclusively collected from ‘\textit{ZiTong}’, spanning 15 major historical periods from the Shang Dynasty (1600 B.C.-1046 B.C.), the earliest Chinese dynasty with written records, to the modern era. This represents the largest Chinese character dataset in terms of historical coverage to our best knowledge. As shown in Fig.~\ref{AS}d, each dynasty category contains no fewer than 2,000 samples, with 9 historical periods exceeding 10,000 images. This subset is divided into 181,187 training samples and 77,643 testing samples.

\subsection{Challenging Tasks}
The MCCD is a multi-attribute, comprehensive dataset featuring a rich variety of annotation types, including character, calligraphy styles, historical dynasties, and famous calligraphers. These detailed attribute information enable the dataset to support a wide range of sophisticated tasks and address various challenges across different research domains. By providing diversity of attributes, MCCD can be adapted to the exploration of various challenging tasks, which are summarized as follows:

\textbf{(1) Calligraphic Chinese Character Recognition Task:} Recognizing Chinese calligraphy characters presents significant challenges due to their intricate stroke connections, diverse script styles, and substantial structural discrepancies across scripts for the same character. Furthermore, even within a single script style, distinct stylistic nuances among calligraphers further complicate accurate recognition of characters, script types, or authorship. MCCD addresses these challenges by providing 7,765 calligraphic character categories spanning 10 script styles from ancient to modern periods. Each category contains no fewer than 8 samples, offering ample and comprehensive data for robust recognition tasks in calligraphic character identification, script classification, and related domains.

\textbf{(2) Calligrapher Identification:} Different calligraphers exhibit distinct stylistic features, but variations in the same character under the same script style are minimal. Additionally, stylistic inconsistencies may occur across different works by the same calligrapher, rendering precise authorship identification highly challenging. To address this, we carefully selected a subset (MCCD-Calligrapher) of 142 famous calligraphers from the MCCD, specifically designed to support research in complex calligraphic writer identification tasks.

\textbf{(3) Multi-Task Learning:} The 258,830 samples from the \textit{‘ZiTong’} source in our dataset include complete annotations of character, calligraphic style, and dynasty. These categories are inherently linked in the historical development of Chinese calligraphy, enabling classifiers to learn interconnected features across their respective tasks. Researchers can leverage this subset to train and evaluate multi-task learning algorithms, while also potentially advancing studies on the evolution of Chinese characters \cite{liu2023frontiers}.

\renewcommand{\arraystretch}{1.25}
\setlength{\tabcolsep}{8pt}
\begin{table}[t]
  \centering
  \caption{Results of character recognition.}
    \resizebox{0.8\textwidth}{!}{
    \begin{tabular}{cccc}
    \toprule
    Method & Top-1 Acc$\uparrow$ & Top-5 Acc$\uparrow$ & Macro Acc$\uparrow$ \\
    \hline
    ResNet50 \cite{he2016deep} & 79.085\% & 93.536\% & 75.598\% \\
    Vision Transformer \cite{dosovitskiy2020image} & 75.352\% & 91.982\% & 70.902\% \\
    Swin Transformer \cite{liu2022swin} & \textbf{79.141\%} & \textbf{93.919\%} & \textbf{75.949\%} \\
    HierCode \cite{zhang2025hiercode} & 60.225\% & 72.113\% & 46.650\% \\
    CCR-CLIP \cite{yu2023chinese} & 67.670\% & 88.570\% & 63.730\% \\
   \bottomrule
    \end{tabular}}
  \label{tab:Character Recognition}%
\end{table}%

\begin{figure}[t]
\centering
\includegraphics[width=1\linewidth]{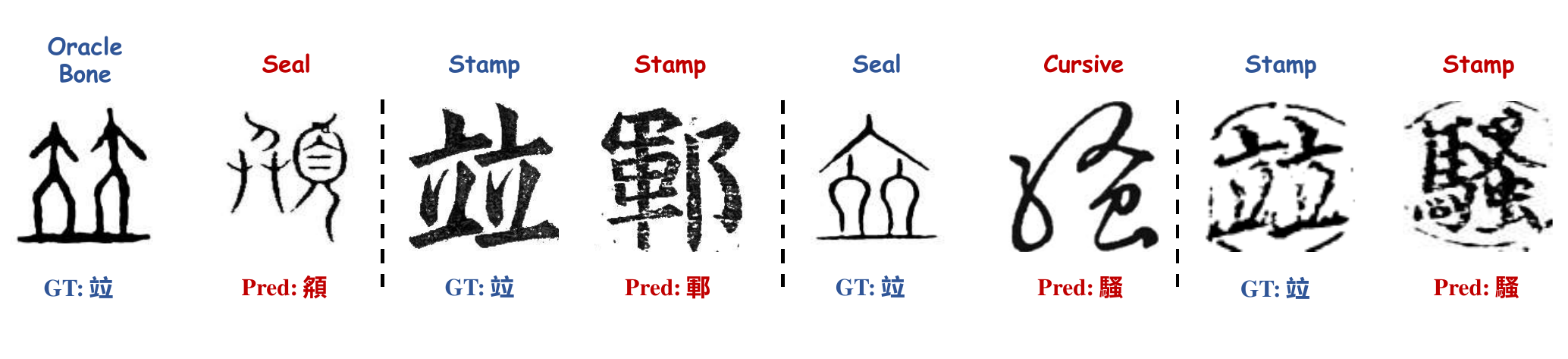}
\caption{Visualization results of character recognition prediction error samples.} \label{CV}
\end{figure}

\section{Experiments}
\subsection{Experiment Setup}
To the best of our knowledge, our dataset is the first open-source Chinese calligraphy character dataset to contain multi-attribute labels. To comprehensively evaluate the performance of the proposed MCCD, we conducted character, script style, dynasty and calligrapher recognition and multi-task learning experiments. All methods were implemented under identical parameter settings to ensure fair and rigorous comparisons. The models were optimized using AdamW \cite{loshchilov2017decoupled} with an initial learning rate of 0.01. Input images were resized to 96×96 pixels, and training data was augmented via RandAugment \cite{cubuk2020randaugment} (severity=9, depth=2, augmentation\_set=‘all’). All models were trained for 100 epochs with a batch size of 128, adopting a MultiStepLR learning rate scheduler that reduced the rate by half every 10 epochs. Evaluation metrics included Top-1 Accuracy, Top-5 Accuracy and Macro Accuracy \cite{chen2021benchmarking} that refers to calculating the accuracy for each class separately and then averaging these per-class accuracies. Experiments were performed on a single NVIDIA GeForce RTX 3090 GPU.

\renewcommand{\arraystretch}{1.25}
\setlength{\tabcolsep}{8pt}
\begin{table}[t]
  \centering
  \caption{Results of calligraphic style recognition.}
    \resizebox{0.8\textwidth}{!}{
    \begin{tabular}{cccc}
    \toprule
    Method & Top-1 Acc$\uparrow$ & Top-5 Acc$\uparrow$ & Macro Acc$\uparrow$ \\
    \hline
    ResNet50 \cite{he2016deep} & \textbf{95.405\%} & 99.995\% & \textbf{94.579\%} \\
    Vision Transformer \cite{dosovitskiy2020image} & 92.588\% & 99.979\% & 91.322\% \\
    Swin Transformer \cite{liu2022swin} & 95.246\% & \textbf{99.996\%} & 94.353\% \\
    \bottomrule
    \end{tabular}}
  \label{tab:Script Style}%
\end{table}%

\begin{figure}[t]
\centering
\includegraphics[width=1\linewidth]{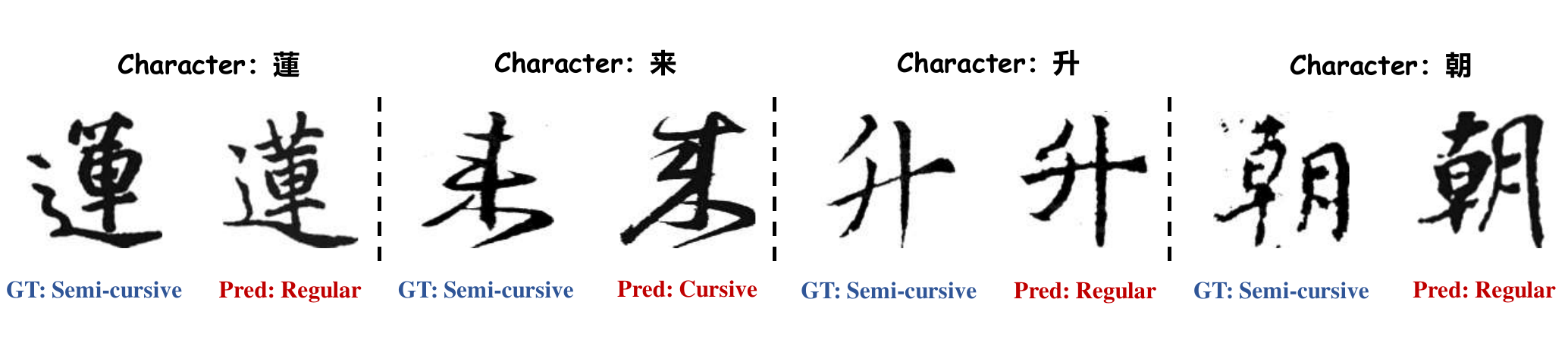}
\caption{Visualization results of calligraphic style recognition prediction error samples.}\label{SV}
\end{figure}

\subsection{Single-task Recognition Experiments}
We conducted recognition experiments on MCCD and all of the three subsets (MCCD-Style, MCCD-Dynasty, MCCD-Calligrapher) to analyze the impact of different backbone networks on classification accuracy. Comparative evaluations were performed using CNN-based architectures (ResNet50 \cite{he2016deep}) and Transformer-based architectures (Vision Transformer \cite{dosovitskiy2020image}, Swin Transformer \cite{liu2022swin}) across all dataset versions. For the extensively studied character recognition task, we additionally implemented SOTA methods: a radical embedding-based approach (HierCode \cite{zhang2025hiercode}) and an image-IDS matching method (CCR-CLIP \cite{yu2023chinese}), both employing ResNet50 \cite{he2016deep} as the backbone network to ensure fair comparisons.

\textbf{Character Recognition:} \textbf{(1) Structural Similarity Analysis.} As shown in Table~\ref{tab:Character Recognition}, most models successfully include the correct category within their Top-5 predictions. However, these methods exhibit a significant gap (about 10\% to 15\%) between Top-1 and Top-5 accuracies. This arises from exaggerated stroke variations or structural omissions caused by cursive connections in calligraphic characters, which amplify ambiguity among visually similar characters. \textbf{(2) Intra-Class Diversity.} There is a notable difference between different calligraphic styles of Chinese calligraphy characters such as the long-established oracle bone script, bronze script, among others. This diversity leads to variations in stroke structure within the same character. As illustrated in Table~\ref{tab:Character Recognition}, the recognition performance of Swin Transformer \cite{liu2022swin} and ResNet50 \cite{he2016deep} is much higher than that of other methods. Conversely, contemporary state-of-the-art Chinese character recognition models, HierCode \cite{zhang2025hiercode} and CCR-CLIP \cite{yu2023chinese}, prioritize the fixed structure of Chinese characters from the radicals, but perform weakly in the domain of calligraphic characters. We visualized some of the incorrectly predicted samples as shown in Fig.~\ref{CV}. We found that the radicals of the same category of characters in different scripts style can vary greatly, thus leading to poor performance of radical-based methods. However, the use of ResNet50 \cite{he2016deep}, ViT \cite{dosovitskiy2020image} and Swin Transformer \cite{liu2022swin}, which focus on global information, can be used to recognize calligraphic characters at the level of overall glyphs, which is more effective in recognizing complex calligraphic characters.

These findings highlight the necessity for models oriented towards the recognition of calligraphic characters with diverse styles to focus more on the acquisition of global features and enhanced generalization, in order to improve the recognition performance compared to the more detailed local radical structure.

\textbf{Calligraphic Style Recognition:} As demonstrated in Table~\ref{tab:Script Style}, all models exhibit near-perfect performance in terms of Top-5 accuracy. However, a substantial gap remains between Top-1 and Top-5 accuracies. This disparity suggests the presence of numerous stylistic similarities in calligraphy, which can lead to frequent misclassifications by the models, as some mispredicted samples shown in Fig.~\ref{SV}. These findings emphasize the inherent challenges of calligraphy style recognition and offer valuable insights to guide future research in this domain.

\renewcommand{\arraystretch}{1.25}
\setlength{\tabcolsep}{8pt}
\begin{table}[t]
  \centering
  \caption{Results of dynasty recognition.}
  \resizebox{0.8\textwidth}{!}{
    \begin{tabular}{cccc}
    \toprule
    Method & Top-1 Acc$\uparrow$ & Top-5 Acc$\uparrow$ & Macro Acc$\uparrow$ \\
    \hline
    ResNet50 \cite{he2016deep} & \textbf{88.469\%} & \textbf{98.881\%} & \textbf{79.477\%} \\
    Vision Transformer \cite{dosovitskiy2020image} & 81.383\% & 97.514\% & 66.425\% \\
    Swin Transformer \cite{liu2022swin} & 87.291\% & 98.787\% & 77.173\% \\
    \bottomrule
    \end{tabular}}
  \label{tab:Dynasty Recognition}%
\end{table}

\begin{figure}[t]
\centering
\includegraphics[width=0.95\linewidth]{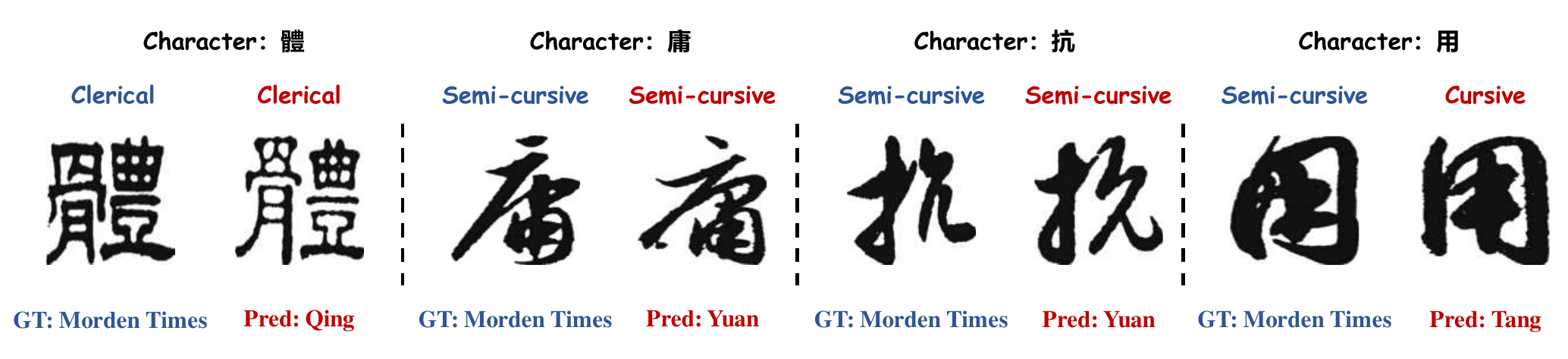}
\caption{Visualization results of dynasty recognition prediction error samples.} \label{DV}
\end{figure}

\textbf{Dynasty Recognition:} \textbf{(1) Overlapping Calligraphic Styles.} As shown in Fig.~\ref{DV}, later historical periods often incorporated and adapted stylistic elements from earlier dynasties, resulting in a blurring of the distinctions between styles and significant ambiguities in recognition. This makes the accurate attribution of calligraphic works to specific periods or styles a challenging task. \textbf{(2) Long-Tailed Distribution.} As shown in Table~\ref{tab:Dynasty Recognition}, all models exhibit a Marco Accuracy lower than Top-1 accuracy, though the gap remains below 15\%. This indicates a mild long-tailed distribution in the MCCD-Dynasty subset, where models struggle to learn discriminative features from tail categories with limited samples.

These challenges mirror real-world difficulties in dynasty recognition of calligraphic works. Our dataset, encompassing comprehensive coverage of major historical periods, provides sufficient data support to address these issues.

\begin{figure}[t]
\centering
\includegraphics[width=0.95\linewidth]{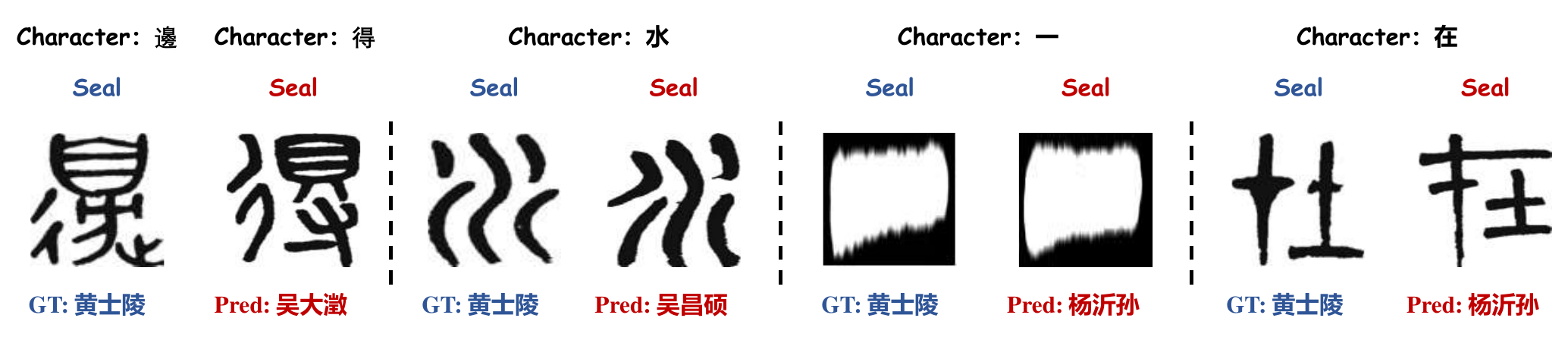}
\caption{Visualization results of calligrapher identification prediction error samples.} \label{mis_sam}
\end{figure}

\textbf{Calligrapher Identification:} From Table~\ref{tab:Calligrapher Identification}, it can be concluded that calligrapher identification proved more challenging compared to the other three recognition tasks, with Top-5 Accuracy peaking at 88.209\%. We conducted a visual analysis of some mispredicted samples in Fig.~\ref{mis_sam}, revealing two key challenges: \textbf{(1) Structural Simplicity.} Characters with simplistic stroke structures (e.g., ‘\begin{CJK*}{UTF8}{gbsn}一\end{CJK*}’ [‘one’]) fail to capture distinctive stylistic features of individual calligraphers. \textbf{(2) Stylistic Homogeneity.} Calligraphers using similar calligraphic styles create overlapping patterns, complicating individual style differentiation. This highlights new challenges for writer identification in our diverse dataset.

\renewcommand{\arraystretch}{1.25}
\setlength{\tabcolsep}{8pt}
\begin{table}[t]
  \centering
  \caption{Results of calligrapher identification.}
  \resizebox{0.8\textwidth}{!}{
    \begin{tabular}{cccc}
    \toprule
    Method & Top-1 Acc$\uparrow$ & Top-5 Acc$\uparrow$ & Macro Acc$\uparrow$ \\
    \hline
    ResNet50 \cite{he2016deep} & \textbf{67.676\%} & \textbf{88.209\%} & \textbf{60.265\%} \\
    Vision Transformer \cite{dosovitskiy2020image} & 41.290\% & 72.170\% & 32.199\% \\
    Swin Transformer \cite{liu2022swin} & 52.614\% & 81.342\% & 44.515\% \\
    \bottomrule
    \end{tabular}}
  \label{tab:Calligrapher Identification}%
\end{table}%

According to the single-task experimental results, ResNet outperforms ViTs on attribute recognition (style, dynasty, calligrapher), likely due to its efficacy on moderately-sized datasets that are challenging for ViTs without extensive pre-training. Conversely, for the high-cardinality task of character recognition, the Swin Transformer \cite{liu2022swin} is better suited. Its shifted window mechanism enables the capture of fine-grained local features, providing a crucial advantage in distinguishing among numerous, visually similar classes.

\subsection{Multi-Task Learning Experiments}
\renewcommand{\arraystretch}{1.25}
\setlength{\tabcolsep}{8pt}
\begin{table}[t]
  \centering
  \caption{Results of dual-task learning. ‘Dataset’ indicates the subset used in the experiment, and ‘Task’ denotes the recognition task.}
    \resizebox{0.8\textwidth}{!}{
    \begin{tabular}{ccccc}
    \toprule
    \multicolumn{1}{c}{Dataset} & Task & Top-1 Acc$\uparrow$ & Top-5 Acc$\uparrow$ & Macro Acc$\uparrow$ \\
    \midrule
    \multirow{2}[4]{*}{MCCD-Style} & Style & 95.327\% & 99.992\% & 94.483\% \\
    \cmidrule(lr){2-5}
          & Character & 76.567\% & 92.359\% & 73.217\% \\
    \midrule
    \multirow{2}[4]{*}{MCCD-Dynasty} & Dynasty & 86.649\% & 98.514\% & 75.752\% \\
    \cmidrule(lr){2-5}
          & Character & 74.376\% & 91.456\% & 69.541\% \\
    \midrule
    \multirow{2}[4]{*}{MCCD-Calligrapher} & Calligrapher & 59.757\% & 84.307\% & 50.035\% \\
    \cmidrule(lr){2-5}
          & Character & 61.403\% & 81.987\% & 47.941\% \\
    \bottomrule
    \end{tabular}}
  \label{tab:Two-Task}%
\end{table}%
Multi-task learning experiments were conducted on each of MCCD's three subsets using their corresponding attribute and character labels for dual-task classification experiments. Furthermore, 67,628 samples containing all four attribute annotations were selected from the calligrapher subset for four-task classification experiments. The objective of this study was to analyze whether multi-task learning can positively impact the recognition performance by learning generalized feature representations. To the best of our knowledge, this is the first study in calligraphic recognition to simultaneously leverage multi-attribute annotations. For fair and direct comparison with single-task baselines, we constructed a foundational multi-task model based on ResNet50 \cite{he2016deep} (chosen for its reliability and strong performance) by modifying the final classification layer of the network from a single classification head to a number that corresponds to the number of tasks.

\textbf{Dual-task Recognition:} The results of the dual-task learning experiment are shown in Table~\ref{tab:Two-Task}. Compared to ResNet50 \cite{he2016deep} single-task baselines, dual-task learning exhibited varying degrees of performance degradation in all metrics. Character recognition performance improved marginally when paired with tasks demonstrating stronger individual performance. Causes are manifold. \textbf{(1) Feature Level Differences.} Character recognition and other tasks (style, dynasty, calligrapher) require distinct feature representations, leading to conflicting optimization objectives. Character recognition needs to capture fine-grained stroke features, while style and dynasty recognition rely more on high-level features such as overall layout, and calligrapher recognition needs to meticulously differentiate the subtle features of individual styles. This difference in feature levels makes muti-task learning difficult. \textbf{(2) Information Leakage Between Tasks.} There may be information leakage or redundancy between certain tasks, for example, specific dynasties may have specific writing styles, which causes the model to overfit these associations instead of learning truly useful features. \textbf{(3) Data Distribution Effects.} The subset of style and dynasty, with their balanced sample distributions, provided limited complementary benefits to character recognition in joint training scenarios.

\textbf{Four-task Recognition:} As shown in Table~\ref{tab:Four-Task}, increasing the number of tasks from two to four (character and three-attribute recognition) leads to a greater decline in performance. The potential reasons for this include: \textbf{(1) Reduced Training Data.} As it was limited to samples with all four labels selected from the MCCD-Calligrapher subset, this reduction in the training set size likely contributes to the diminished recognition performance. \textbf{(2) Increased Task Interference.} The increased number of tasks (character, calligrapher, style, dynasty) led to greater conflict and competition for feature extraction. \textbf{(3) Imbalanced Learning Difficulty.} The character recognition task is significantly harder to optimize within the multi-task learning framework due to its vastly larger number of classes compared to the other attribute tasks.

Our analysis reveals that the inherent disparities in the quantity of calligraphic works across dynastic periods and calligraphers are naturally reflected in our dataset, which aligns with real-world distributions while offering diverse annotation types. In the future, by fine-tuning the model algorithms and optimizing the shared-layer architecture, we can explore the feature relationships among different attributes of calligraphic character in greater depth, alleviate the challenge of unbalanced sample distribution, and promote research on the evolutionary trajectory of Chinese calligraphy.

\begin{table}[t]
  \centering
  \caption{Results of four-task learning. ‘Dataset’ indicates the subset used in the experiment, and ‘Task’ denotes the recognition task.}
    \resizebox{0.8\textwidth}{!}{
    \begin{tabular}{ccccc}
    \toprule
    \multicolumn{1}{c}{Dataset} & Task & Top-1 Acc$\uparrow$ & Top-5 Acc$\uparrow$ & Macro Acc$\uparrow$ \\
    \midrule
    \multirow{4}[4]{*}{MCCD-Calligrapher} & Calligrapher & 56.671\% & 82.146\% & 45.412\% \\
    \cmidrule(lr){2-5}
          & Character & 52.792\% & 75.702\% & 37.879\% \\
    \cmidrule(lr){2-5}
          & Style & 87.147\% & 100.0\% & 90.101\% \\
    \cmidrule(lr){2-5}
          & Dynasty & 71.526\% & 97.139\% & 61.058\% \\
    \bottomrule
    \end{tabular}}
  \label{tab:Four-Task}%
  \vspace{-0.5cm}
\end{table}%


\section{Conclusion}
In this paper, we present a novel Multi-attribute Chinese Calligraphy Character Dataset (MCCD), comprising 329,715 image samples. All the samples in MCCD are character-labeled, covering a total of 7,765 classes of Chinese characters. In addition, we constructed three subsets from each sample based on the attribute information of the Chinese calligraphy characters present in each sample: (1) MCCD-Style: a subset of calligraphic style attribute covering 10 calligraphy types, (2) MCCD-Dynasty: a subset of dynasty attribute spanning 15 historical periods, and (4) MCCD-Calligrapher: a subset of calligrapher attribute featuring 142 renowned calligraphers based on the corresponding attribute information. With its comprehensive annotations and abundant samples, MCCD enables diverse research tasks including character, style, and dynasty recognition of Chinese calligraphy characters, as well as writer identification, multitask learning, and more. Based on this dataset, we provide benchmark results of recognition experiments for all attributes of the dataset, and multi-task recognition experiments combining characters with the other three types of labeled information. The experimental results demonstrate that a dataset with systematically diverse attribute labeling and sufficient samples can provide substantial data support for different kinds of research. At the same time, given the the complexity of the stroke structure and writing style of the Chinese Calligraphy characters, this multi-attribute annotated calligraphy dataset poses new challenges for the field of Chinese character recognition, which need to be explored even further by future methods. It is anticipated that this dataset will enable researchers to undertake in-depth studies of Chinese calligraphy characters from a broad range of perspectives.

\section*{Acknowledgement}
This research is supported in part by the National Natural Science Foundation of China (Grant No.: 62476093), and the National Key Research and Development Program of China (2022YFC3301703).
%
%
%

\begin{thebibliography}{10}
\providecommand{\url}[1]{\texttt{#1}}
\providecommand{\urlprefix}{URL }
\providecommand{\doi}[1]{https://doi.org/#1}

\bibitem{chen2021benchmarking}
Chen, J., Yu, H., Ma, J., Guan, M., Xu, X., Wang, X., Qu, S., Li, B., Xue, X.: Benchmarking chinese text recognition: Datasets, baselines, and an empirical study. arXiv preprint arXiv:2112.15093  \textbf{3}(4), ~5 (2021)

\bibitem{cubuk2020randaugment}
Cubuk, E.D., Zoph, B., Shlens, J., Le, Q.V.: Randaugment: Practical automated data augmentation with a reduced search space. In: Proceedings of the IEEE/CVF conference on computer vision and pattern recognition workshops. pp. 702--703 (2020)

\bibitem{dosovitskiy2021an}
Dosovitskiy, A., Beyer, L., Kolesnikov, A., Weissenborn, D., Zhai, X., Unterthiner, T., Dehghani, M., Minderer, M., Heigold, G., Gelly, S., Uszkoreit, J., Houlsby, N.: An image is worth 16x16 words: Transformers for image recognition at scale. In: International Conference on Learning Representations (2021)

\bibitem{guan2024open}
Guan, H., Wan, J., Liu, Y., Wang, P., Zhang, K., Kuang, Z., Wang, X., Bai, X., Jin, L.: An open dataset for the evolution of oracle bone characters: Evobc. arXiv preprint arXiv:2401.12467  (2024)

\bibitem{he2016deep}
He, K., Zhang, X., Ren, S., Sun, J.: Deep residual learning for image recognition. In: Proceedings of the IEEE conference on computer vision and pattern recognition. pp. 770--778 (2016)

\bibitem{jin2011scut}
Jin, L., Gao, Y., Liu, G., Li, Y., Ding, K.: Scut-couch2009—a comprehensive online unconstrained chinese handwriting database and benchmark evaluation. International Journal on Document Analysis and Recognition (IJDAR)  \textbf{14},  53--64 (2011)

\bibitem{li2022swordnet}
Li, X., Wang, J., Zhang, H., Huang, Y., Huang, H.: Swordnet: Chinese character font style recognition network. IEEE Access  \textbf{10},  8388--8398 (2022)

\bibitem{li2008scut}
Li, Y., Jin, L., Zhu, X., Long, T.: Scut-couch2008: A comprehensive online unconstrained chinese handwriting dataset. ICFHR  \textbf{2008},  165--170 (2008)

\bibitem{liu2023frontiers}
Liu, C., Jin, L., Bai, X., Li, X., Yin, F.: Frontiers of intelligent document analysis and recognition: review and prospects. Journal of Image and Graphics  \textbf{28}(08),  2223--2252 (2023)

\bibitem{liu2011casia}
Liu, C.L., Yin, F., Wang, D.H., Wang, Q.F.: Casia online and offline chinese handwriting databases. In: 2011 international conference on document analysis and recognition. pp. 37--41. IEEE (2011)

\bibitem{liu2022swin}
Liu, Z., Hu, H., Lin, Y., Yao, Z., Xie, Z., Wei, Y., Ning, J., Cao, Y., Zhang, Z., Dong, L., et~al.: Swin transformer v2: Scaling up capacity and resolution. In: Proceedings of the IEEE/CVF conference on computer vision and pattern recognition. pp. 12009--12019 (2022)

\bibitem{loshchilov2018decoupled}
Loshchilov, I., Hutter, F.: Decoupled weight decay regularization. In: International Conference on Learning Representations (2019)

\bibitem{ma2020joint}
Ma, W., Zhang, H., Jin, L., Wu, S., Wang, J., Wang, Y.: Joint layout analysis, character detection and recognition for historical document digitization. In: 2020 17th International Conference on Frontiers in Handwriting Recognition (ICFHR). pp. 31--36. IEEE (2020)

\bibitem{pengcheng2017chinese}
Pengcheng, G., Gang, G., Jiangqin, W., Baogang, W.: Chinese calligraphic style representation for recognition. International Journal on Document Analysis and Recognition (IJDAR)  \textbf{20},  59--68 (2017)

\bibitem{shi2025large}
Shi, Y., Peng, D., Zhang, Y., Cao, J., Jin, L.: A large-scale dataset for chinese historical document recognition and analysis. Scientific Data  \textbf{12}(1), ~169 (2025)

\bibitem{su2006hit}
Su, T., Zhang, T., Guan, D.: Hit-mw dataset for offline chinese handwritten text recognition. In: Tenth International Workshop on Frontiers in Handwriting Recognition. Suvisoft (2006)

\bibitem{wang2024puzzle}
Wang, P., Zhang, K., Wang, X., Han, S., Liu, Y., Jin, L., Bai, X., Liu, Y.: Puzzle pieces picker: Deciphering ancient chinese characters with radical reconstruction. In: International Conference on Document Analysis and Recognition. pp. 169--187. Springer (2024)

\bibitem{wang2024open}
Wang, P., Zhang, K., Wang, X., Han, S., Liu, Y., Wan, J., Guan, H., Kuang, Z., Jin, L., Bai, X., et~al.: An open dataset for oracle bone character recognition and decipherment. Scientific Data  \textbf{11}(1), ~976 (2024)

\bibitem{xu2019casia}
Xu, Y., Yin, F., Wang, D.H., Zhang, X.Y., Zhang, Z., Liu, C.L.: Casia-ahcdb: A large-scale chinese ancient handwritten characters database. In: 2019 International Conference on Document Analysis and Recognition (ICDAR). pp. 793--798. IEEE (2019)

\bibitem{Yang2023Open}
Yang, C., Liu, C., Fang, Z., Han, Z., Liu, C., Yin, X.: Open set text recognition technology. Journal of Image and Graphics  \textbf{28}(6),  1767--1791 (2023)

\bibitem{yang2018dense}
Yang, H., Jin, L., Huang, W., Yang, Z., Lai, S., Sun, J.: Dense and tight detection of chinese characters in historical documents: Datasets and a recognition guided detector. IEEE Access  \textbf{6},  30174--30183 (2018)

\bibitem{yu2023chinese}
Yu, H., Wang, X., Li, B., Xue, X.: Chinese text recognition with a pre-trained clip-like model through image-ids aligning. In: Proceedings of the IEEE/CVF International Conference on Computer Vision. pp. 11943--11952 (2023)

\bibitem{zhang2009hcl2000}
Zhang, H., Guo, J., Chen, G., Li, C.: Hcl2000-a large-scale handwritten chinese character database for handwritten character recognition. In: 2009 10th international conference on document analysis and recognition. pp. 286--290. IEEE (2009)

\bibitem{zhang2025megahan97k}
Zhang, Y., Shi, Y., Zhang, P., Zhao, Y., Yang, Z., Jin, L.: Megahan97k: A large-scale dataset for mega-category chinese character recognition with over 97k categories. Pattern Recognition p. 111757 (2025)

\bibitem{zhang2025hiercode}
Zhang, Y., Zhu, Y., Peng, D., Zhang, P., Yang, Z., Yang, Z., Yao, C., Jin, L.: Hiercode: A lightweight hierarchical codebook for zero-shot chinese text recognition. Pattern Recognition  \textbf{158},  110963 (2025)

\bibitem{zhou2010hit}
Zhou, S., Chen, Q., Wang, X.: Hit-or3c: an opening recognition corpus for chinese characters. In: Proceedings of the 9th IAPR International Workshop on document analysis systems. pp. 223--230 (2010)

\end{thebibliography}
%

\end{document}